\documentclass[10pt,twocolumn,letterpaper]{article}

\usepackage{cvpr}
\usepackage{times}
\usepackage{epsfig}
\usepackage{graphicx}
\usepackage{amsmath}
\usepackage{amssymb}

\long\def\ignorethis#1{}

\usepackage{amsmath}
\usepackage{mathptmx}
\usepackage{floatrow}
\usepackage{xfrac}

\usepackage{color}
\definecolor{gray}{rgb}{0.35,0.35,0.35}
\definecolor{green}{rgb}{0,1,0.2}
\definecolor{blue}{rgb}{0,0,1}
\definecolor{white}{rgb}{1,1,1}

\usepackage{xcolor}

\usepackage{epsfig}
\usepackage{epstopdf}
\usepackage{subfig}
\usepackage{multirow}
\usepackage{overpic}
\usepackage{array}
\usepackage{booktabs}
\usepackage{calrsfs}
\DeclareMathAlphabet{\pazocal}{OMS}{zplm}{m}{n}

\newcommand\blfootnote[1]{
	\begingroup
	\renewcommand\thefootnote{}\footnote{#1}
	\addtocounter{footnote}{-1}
	\endgroup
}

\usepackage[pagebackref=true,breaklinks=true,letterpaper=true,colorlinks,bookmarks=false]{hyperref}

\cvprfinalcopy 

\def\cvprPaperID{4824}

\usepackage{titling}
\date{}

\begin{document}

\title{\Large{\textbf{ScrabbleGAN: Semi-Supervised Varying Length Handwritten Text Generation}}}

\author{
Sharon Fogel$^\dagger$, Hadar Averbuch-Elor$^\mathsection$, Sarel Cohen$^\dagger$, Shai Mazor$^\dagger$ and Roee Litman$^\dagger$ \\
$^\dagger$ Amazon Rekognition, Israel \qquad $^\mathsection$ Cornell Tech, Cornell University
}

\maketitle

\begin{abstract}

Optical character recognition (OCR) systems performance have improved significantly in the deep learning era.
This is especially true for handwritten text recognition (HTR), where each author has a unique style, unlike printed text, where the variation is smaller by design.
That said, deep learning based HTR is limited, as in every other task, by the number of training examples.
Gathering data is a challenging and costly task, and even more so, the labeling task that follows, of which we focus here.
One possible approach to reduce the burden of data annotation is semi-supervised learning.
Semi supervised methods use, in addition to labeled data, some unlabeled samples to improve performance, compared to fully supervised ones.
Consequently, such methods may adapt to unseen images during test time.

We present ScrabbleGAN, a semi-supervised approach to synthesize handwritten text images that are versatile both in style and lexicon.
ScrabbleGAN relies on a novel generative model which can generate images of words with an arbitrary length.
We show how to operate our approach in a semi-supervised manner, enjoying the aforementioned benefits such as performance boost over state of the art supervised HTR.
Furthermore, our generator can manipulate the resulting text style.
This allows us to change, for instance, whether the text is cursive, or how thin is the pen stroke.

\end{abstract}
\begin{figure}
	\includegraphics[width=\columnwidth]{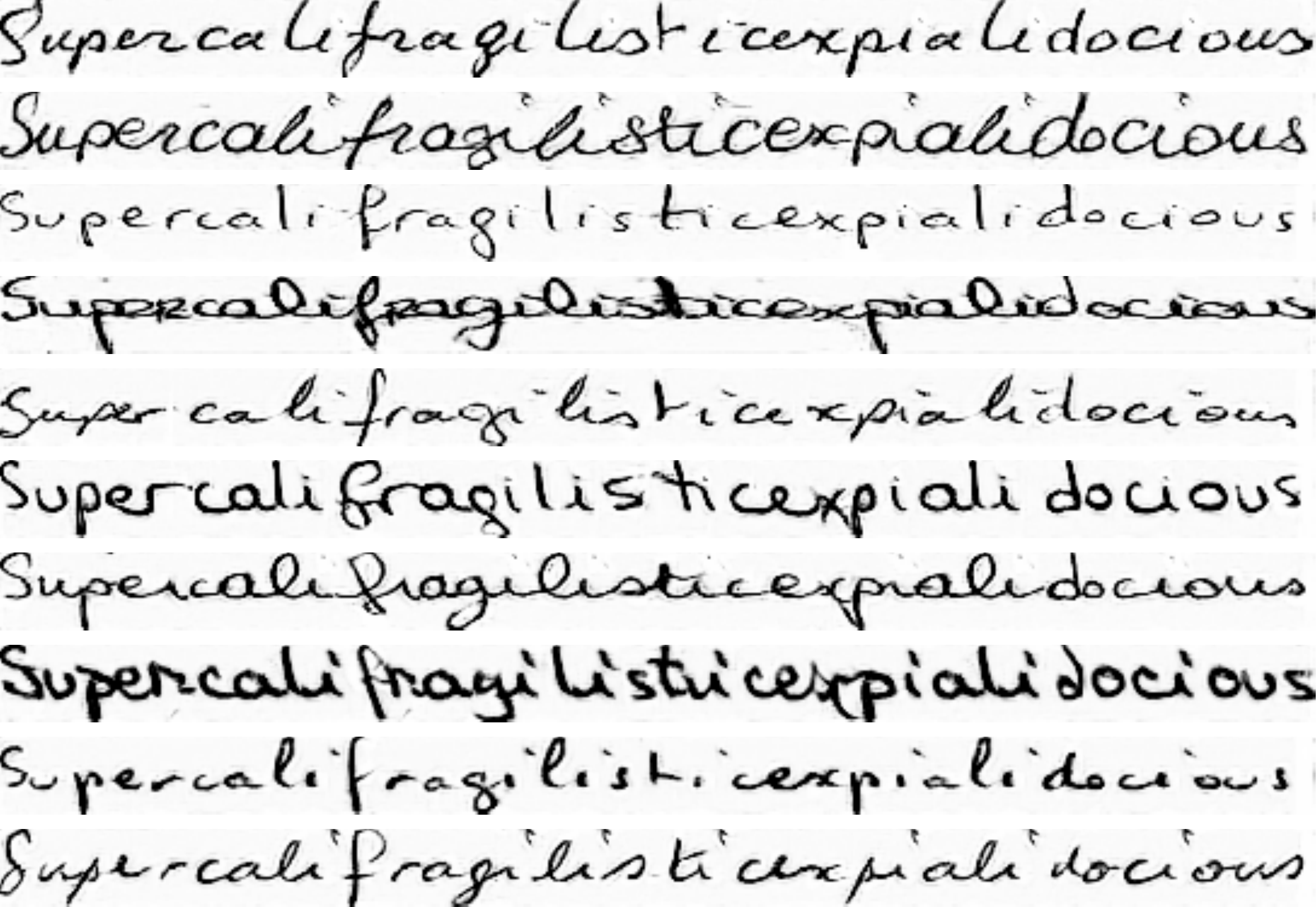}
	\caption{\label{fig:supercali}
		The  word ``Supercalifragilisticexpialidocious'' (34 letters) from the movie ``Mary Poppins'' written in different styles using our network. Note that some of these styles are cursive.
		\vspace{-2ex}
	}	
\end{figure}

\blfootnote{Corresponding author:  \texttt{shafog@amazon.com}}
\blfootnote{$^\mathsection$ work done while working at Amazon.}

\section{Introduction}

Documentation of knowledge using handwriting is one of the biggest achievements of mankind: the oldest written records mark the transition from prehistory into history, and indeed, most evidence of historic events can be found in handwritten scripts and markings.
Handwriting remained the dominant way of documenting events and data well after Gutenberg's printing press in the mid-1400s.
Both printing and handwriting are becoming somewhat obsolete in the digital era, when courtroom stenographers are being replaced by technology \cite{bbc}, further, most of the text we type remains in digital form and never meets a paper.

Nevertheless, handwritten text still has many applications today, a huge of amount of handwritten text has accumulated over the years, ripe to be processed, and still continues to be written today.
Two prominent cases where handwriting is still being used today are healthcare and financial institutions. 
There is a growing need for those to be extracted and made accessible, e.g. by modern search engines.  
While modern OCRs seem to be mature enough to handle printed text \cite{textract, gcp_ocr}, \emph{handwritten text recognition} (HTR) does not seem to be on par. 
We attribute this gap to both the lack of versatile, annotated handwritten text, and the difficulty to obtain it. 
In this work, we attempt to address this gap by creating real-looking synthesized text, reducing the need for annotations and enriching the variety of training data in both style and lexicon.

\begin{figure*}[t]
 	\begin{overpic}[width=.9\textwidth]{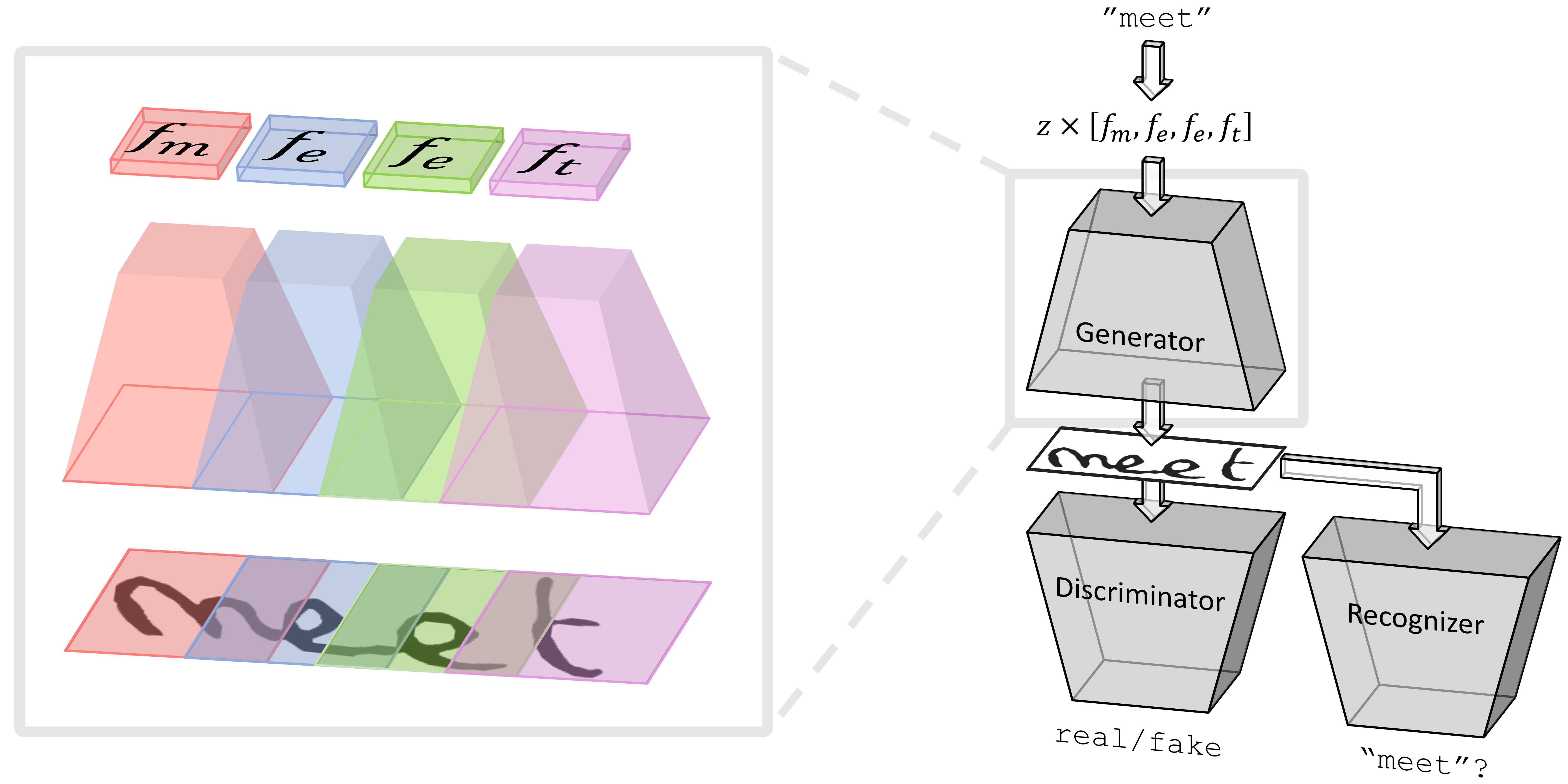}
 		\put(72,31){\LARGE $\pazocal{G}$}
 		\put(71,6.5){\LARGE $\pazocal{D}$}
 		\put(89,05){\LARGE $\pazocal{R}$}
 	\end{overpic}
	\caption{\label{fig:arch}
		{\em Architecture overview} for the case of generating the word ``meet''.
		\textbf{Right}: Illustration of the entire ScrabbleGAN architecture.		
		Four character filters are concatenated ($f_e$ is used twice), multiplied by the noise vector $z$ and fed into the generator $\pazocal{G}$. 
		The resulting image is fed into both the discriminator  $\pazocal{D}$ and the recognizer $\pazocal{R}$ , respectively promoting style and data fidelity.
		\textbf{Left}: A detailed illustration of the generator network $\pazocal{G}$, showing how the concatenated filters are each fed into a class-conditioned generator, where the resulting receptive fields thereof are overlapping. This overlap allows for adjacent characters to interact, enabling cursive text, for example.
		}	
\end{figure*}

\paragraph{Our contributions} are threefold;
First, we present a novel fully convolutional handwritten text generation architecture, which allows for arbitrarily long outputs.
This is in contrast to the vast majority of text related solutions which rely on recurrent neural networks (RNN).
Our approach is able to generate arbitrarily long words (e.g., see Figure~\ref{fig:supercali}) or even complete sentences altogether.
Another benefit of this architecture is that it learns character embeddings without the need for character level annotation.
Our method's name was chosen as an analogy between the generation process to the way words are created during the game of Scrabble, i.e. by \emph{concatenating} some letter-tokens together into a word.
Second, we show how to train this generator in a semi-supervised regime, allowing adaptation to unlabeled data in general, and specifically to the test time images. 
To the best of our knowledge, this is the first use of {\em unlabeled} data to train a handwritten text synthesis framework.
Finally, we provide empirical evidence that the training lexicon matters no less than the richness of styles for HTR training. 
This fact emphasizes the advantage of our method over ones that only warp and manipulate the training images.

\section{Previous Work}

\paragraph{Handwriting text recognition} can be seen as a specific case of \emph{optical character recognition} (OCR). 
This is a well studied topic, in the in-depth survey \cite{plamondon2000online}, HTR approaches are divided into online and offline methods, which differ by the type of data they consume:
Online methods have access to the pen location as the text is being written, and hence can disambiguate intersecting strokes. 
Offline methods, conversely, have access only to the \emph{\textit}{final} resulting text image (i.e. rasterized), possibly also in the presence of some background noise or clutter. 
Clearly, online methods have a strict advantage over their offline counterparts in terms of data quality, but require additional equipment (such as a touchscreen) to capture pen stroke data.
Hence, online data is harder to create in large quantities, especially in a natural setting. 
Furthermore, these methods are unsuitable for historic manuscripts and markings which are entirely offline.
For this reason, we chose to focus on offline methods and leave online methods out of the scope for this manuscript.

Modern HTR methods harness the recent advancements in deep networks, achieving top performance on most, if not all, modern benchmarks. 
Many of these methods are inspired by the \emph{convolutional recurrent neural network} (CRNN) architecture, used originally for scene text recognition by Shi et al. \cite{shi2016end}.
Poznanski et al. \cite{poznanski2016cnn} used a CNN to estimate the n-grams profile of an image and match it to the profile of an existing word from a dictionary. 
PHOCNet by Sudholt et al. \cite{sudholt2016phocnet} extended the latter by employing a pyramidal histogram of characters (PHOC), which was used mainly for word spotting. 
Suerias et al. \cite{sueiras2018offline} used an architecture inspired by sequence to sequence \cite{sutskever2014sequence}, in which they use an attention decoder rather than using the CRNN outputs directly.
Dutta et al. \cite{dutta2018improving} compiled several recent advances in text recognition into a powerful architecture, reminiscent of modern networks for scene text recognition, as the ones presented recently by Baek et al. \cite{baek2019wrong}.

\paragraph{Handwriting text generation} (HTG) is a relatively new field, brought forth by Graves \cite{graves2013generating}, who introduced a method to synthesize online data based on a recurrent net. 
A modern extension of \cite{graves2013generating} was presented by Ji et al. \cite{ji2019generative}, who followed the GAN paradigm \cite{goodfellow2014generative} by adding a discriminator.
DeepWriting \cite{aksan2018deepwriting} introduced better control over the style generation of \cite{graves2013generating} by disentangling it from the content.

Haines et al. \cite{haines2016my} proposed a method to generate handwriting based on a specific author, but requires a time consuming character-level annotation process for each new data sample.

While all previous HTG methods demonstrate visually pleasing results, none were used to augment HTR training data, as opposed to the ones we discuss next.

\paragraph{Data augmentation using generative models.}  
Generative models (and specifically GANs) are used to synthesize realistic data samples based on real examples. 
One possible use for these newly generated images is adding them to the original training set, essentially augmenting the set in a bootstrap manner.
A recent example for this is the low-shot learning method by Wang et al. \cite{wang2018low}, who incorporate this process into the task loss in an end-to-end manner.

For the case at hand, we look at methods that use HTG or similar approaches to learn augmentation of the handwritten examples.
One straightforward example of this is a method proposed by Bhunia et al. \cite{bhunia2018handwriting}, who trains a GAN to warp the training set using a parametric function. Unlike ours, this approach cannot generate words outside a given lexicon, which is a crucial property as we show below (see Table~\ref{tab:transfer_CVL}). 
Krishanan el al. \cite{krishnan2018word} proposed a method to harness synthetic data for word spotting, while not relying on a specific source of synthetic data (e.g. can use data made by our method).

Alonso et al. \cite{alonso2019adversarial} presented a new HTG model reminiscent of the work in \cite{wang2018low}, which in turn inspired our approach.
The network presented in \cite{alonso2019adversarial} uses LSTM to embed the input word into a fixed length representation which can be fed into a BigGAN \cite{brock2018large} architecture.
As opposed to our approach, which allows for variable word and image length, this generator is only able to output images of a fixed width across all word lengths.
Another large benefit of using a fully convolutional generator is removing the need to learn an embedding of the entire word using a recurrent network, we instead can learn the embeddings for each character directly without the need for character level annotation.

Another recent approach by Ingle et al. \cite{ingle2019scalable} uses an online generator similar to \cite{graves2013generating}, followed by rendering. This approach is coupled with some synthetic generation of noise or other nuisance factors. Since this method relies on an online data generator, it cannot adapt to the versatility nor typical noise of an unseen \emph{offline} dataset, which we claim is the common use case.

\paragraph{Classic augmentation} is mentioned here mainly for completeness, including some methods that use less intricate ways to synthesize training examples, such as using handwriting fonts as proposed by \cite{krishnan2016generating}. 
Most of HTR methods mentioned above use some kind of randomized parametric spatial distortion to enlarge the visual variability of the data.
Puigcerver \cite{puigcerver2017multidimensional} pushed this notion even further, and promoted that simpler one dimensional recurrent layers might be sufficient, if provided with data distortions. 

\section{Method}

Our approach follows the GAN paradigm \cite{goodfellow2014generative}, where in addition to the discriminator $\pazocal{D}$, the resulting image is also evaluated by a text recognition network $\pazocal{R}$.
While $\pazocal{D}$ promotes realistic looking handwriting styles, $\pazocal{R}$ encourages the result to be readable and true to the input text.
This part of our architecture is similar to the one presented in \cite{alonso2019adversarial}, and is illustrated in the right side of Figure~\ref{fig:arch}. This architecture minimizes a joint loss term $\ell$ from the two networks
\begin{equation}\label{eq:loss}
	\ell = \ell_\pazocal{D} + \lambda \cdot \ell_\pazocal{R},
\end{equation}
where $\ell_\pazocal{D}$ and $\ell_\pazocal{R}$ are the loss terms of $\pazocal{D}$ and $\pazocal{R}$, respectively.

The main technical novelty of our method lies in the generator $\pazocal{G}$, as we describe next in Section~\ref{sec:gen}. 
Other modifications made to the discriminator $\pazocal{D}$ and the recognizer $\pazocal{R}$ are covered in sections~\ref{sec:disc} and~\ref{sec:recog}, respectively.
We conclude by covering some optimization considerations on the parameter $\lambda$ in Section~\ref{sec:opt}.

\subsection{Fully convolutional generator}
\label{sec:gen}

The main observation guiding our design is that handwriting is a local process, i.e. when writing each letter is influenced only by its predecessor and successor. 
Evidence for this observation can be seen in previous works like \cite{graves2013generating}, where the attention of the synthesizer is focused on the immediate neighbors of the current letter. 
This phenomenon is not trivial since the architecture in \cite{graves2013generating} uses a recurrent network, which we argue enforces no such constraint on the attention, but is rather `free' to learn it.

Our generator is designed to mimic this process: rather than generating the image out of an entire word representation, as done in \cite{alonso2019adversarial}, each character is generated individually, using CNN's property of overlapping receptive fields to account for the influence of nearby letters. 
In other words, our generator can be seen as a concatenation of \emph{identical} class conditional generators \cite{mirza2014conditional} for which each class is a character.
Each of these generators produces a patch containing its input character.
Each convolutional-upsampling layer widens the receptive field, as well as the overlap between two neighboring characters.
This overlap allows adjacent characters to interact, and creates a smooth transition. 

The generation process is illustrated on the left side of Figure~\ref{fig:arch} for the word ``meet''. 
For each character, a filter $f_\star$ is selected from a filter-bank $\mathcal F$ that is as large as the alphabet, for example $\mathcal F=\{f_a, f_b, \ldots, f_z\}$ for lowercase English.
Four such filters are concatenated in Figure~\ref{fig:arch} ($f_e$ is used twice), and multiplied by a noise vector $z$, which controls the text style.
As can be seen, the region generated from each character filter $f_\star$ is of the same size, and adjacent characters' receptive field overlap.
This provides flexibility in the actual size and cursive type of the output handwriting character.
For example, the letter ``m'' takes up most of the red patch, while the letters ``e'' and ``t'' take up a smaller portion of their designated patches, and the latter is the only non-cursive letter.
Furthermore, learning the dependencies between adjacent characters allows the network to create different variations of the same character, depending on its neighboring characters. Such examples can be seen in Figure~\ref{fig:supercali} and Figure~\ref{fig:styles}.

The style of each image is controlled by a noise vector $z$ given as input to the network.
In order to generate the same style for the entire word or sentence, this noise vector is kept constant throughout the generation of all the characters in the input.

\subsection{Style-promoting discriminator}
\label{sec:disc}

In the GAN paradigm \cite{goodfellow2014generative}, the purpose of the discriminator $\pazocal{D}$ is to tell apart synthetic images generated by  $\pazocal{G}$ from the real ones.
In our proposed architecture, the role of $\pazocal{D}$ is also to discriminate between such images based on the handwriting output \emph{style}.

The discriminator architecture has to account for the varying length of the generated image, and therefore is designed to be convolutional as well:
The discriminator is essentially a concatenation of separate ``real/fake'' classifiers with overlapping receptive fields. 
Since we chose not to rely on character level annotations, we cannot use class supervision for each of these classifiers, as opposed to class conditional GANs such as \cite{mirza2014conditional,brock2018large}.  
One benefit of this is that we can now use unlabeled images to train $\pazocal{D}$, even from other unseen data corpus.
A pooling layer aggregates scores from all classifiers into the final discriminator output.

\subsection{Localized text recognizer}
\label{sec:recog}

While discriminator $\pazocal{D}$ promotes real-looking images, the recognizer $\pazocal{R}$ promotes readable text, in essence discriminating between gibberish and real text.
Generated images are `penalized' by comparing the recognized text in the output of $\pazocal{R}$ to the one that was given as input to  $\pazocal{G}$.
Following \cite{alonso2019adversarial}, $\pazocal{R}$ is trained only on real, labeled, handwritten samples.

Most recognition networks use a recurrent module, typically bidirectional LSTM \cite{hochreiter1997long}, which reads the character in the current image patch by utilizing information from previous and subsequent image patches.
As shown by Sabir el al. \cite{sabir2017implicit}, the network learns an implicit language model which helps it identify the correct character even if it is not written clearly, by leveraging priors learned from other characters in the text.
While this quality is usually desired in a handwriting recognition model, in our case it may lead the network to correctly read characters which were not written clearly by the generator. 
Therefore, we opted not to use the recurrent `head' of the recognition network, which enables this quality, and keep only the convolutional backbone.
See the supplementary material for a detailed analysis on this.

\subsection{Optimization considerations}
\label{sec:opt}
The generator network is optimized by the recognizer loss $\ell_\pazocal{R}$ and the adversarial loss $\ell_\pazocal{D}$.
The gradients stemming from each of these loss terms can vary greatly in magnitude.
Alonso et al. \cite{alonso2019adversarial} proposed the following rule to balance the two loss terms

\begin{equation}
    \nabla_I \pazocal{R} \leftarrow \alpha
    \left(
    \frac{\sigma (\nabla_I \pazocal{D})}{\sigma (\nabla_I \pazocal{R})} \cdot
    [ \nabla_I \pazocal{R} - \mu (\nabla_I \pazocal{R})] + \mu (\nabla_I \pazocal{D} )
    \right),
\label{eq:GB_alfonso}
\end{equation}

where $\sigma(\cdot)$ and $\mu (\cdot) $ are respectively the empirical standard deviation and mean, $\nabla_I \pazocal{R}$ and $\nabla_I \pazocal{D}$ are respectively the gradients of $\ell_\pazocal{R}$ and $\ell_\pazocal{D}$ w.r.t. the image.
The parameter $\alpha$ controls the relative importance of $\ell_\pazocal{R}$ compared to $\ell_\pazocal{D}$.
In this paper, we chose to balance based only on the standard deviation of the losses and not the average
\begin{equation}
    \nabla_I  \pazocal{R} \leftarrow \alpha\left( \frac{\sigma (\nabla_I \pazocal{D})}{\sigma (\nabla_I \pazocal{R})} \cdot \nabla_I  \pazocal{R} \right),
\label{eq:GB_ours}
\end{equation}
in order to avoid changing the sign of the gradient $\nabla_I \pazocal{R}$.

\section{Results}

\subsection{Implementation details}
\label{sec:implementation}
Without loss of generality, the architecture is designed to generate and process images with fixed height of 32 pixels, in addition, the receptive field width of $\pazocal{G}$ is set to 16 pixels.

\begin{figure*}[t]

	\includegraphics[width=\textwidth]{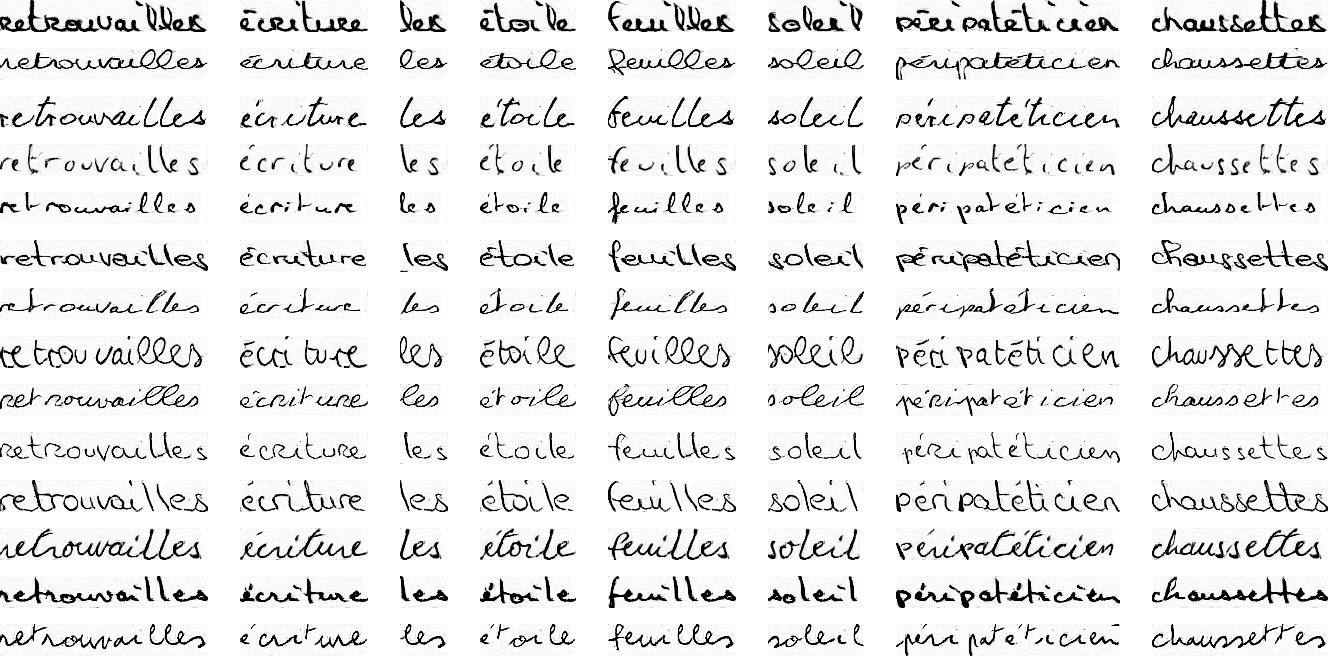}
	\caption{\label{fig:styles}
		{\em Generating different styles.}
		Each row in the figure is generated by the same noise vector and results in the same handwriting style. 
		The words generated in each column from left to right are: retrouvailles, \'ecriture, les, \'etoile, feuilles, soleil, p\'eripat\'eticien and chaussettes
	}
\end{figure*}

As mentioned in Section~\ref{sec:gen}, the generator network $\pazocal{G}$ has a filter bank $\mathcal F$ as large as the alphabet, for example, $\mathcal F=\{f_a, f_b, \ldots, f_z\}$ for lowercase English.
Each filter has a size of $32 \times 8192$.
To generate one $n$-character word, we select and concatenate $n$ of these filters (including repetitions, as with the letter `e' in Figure~\ref{fig:arch}), multiplying them with a $32$ dimensional noise vector $z_1$, resulting in an $n \times 8192$ matrix. 
Next, the latter matrix is reshaped into a $512 \times 4 \times 4n$ tensor, i.e. at this point, each character has a spatial size of $4\times 4$.
The latter tensor is fed into three residual blocks which upsample the spatial resolution, create the aforementioned receptive field overlap, and lead to the final image size of $32 \times 16n$.
Conditional Instance Normalization layers \cite{dumoulin2016learned} are used to modulate the residual blocks using three additional $32$ dimensional noise vectors, $z_2, z_3$ and $z_4$.
Finally, a convolutional layer with a $tanh$ activation is used to output the final image.

\begin{figure}[t]
	\includegraphics[width=\columnwidth]{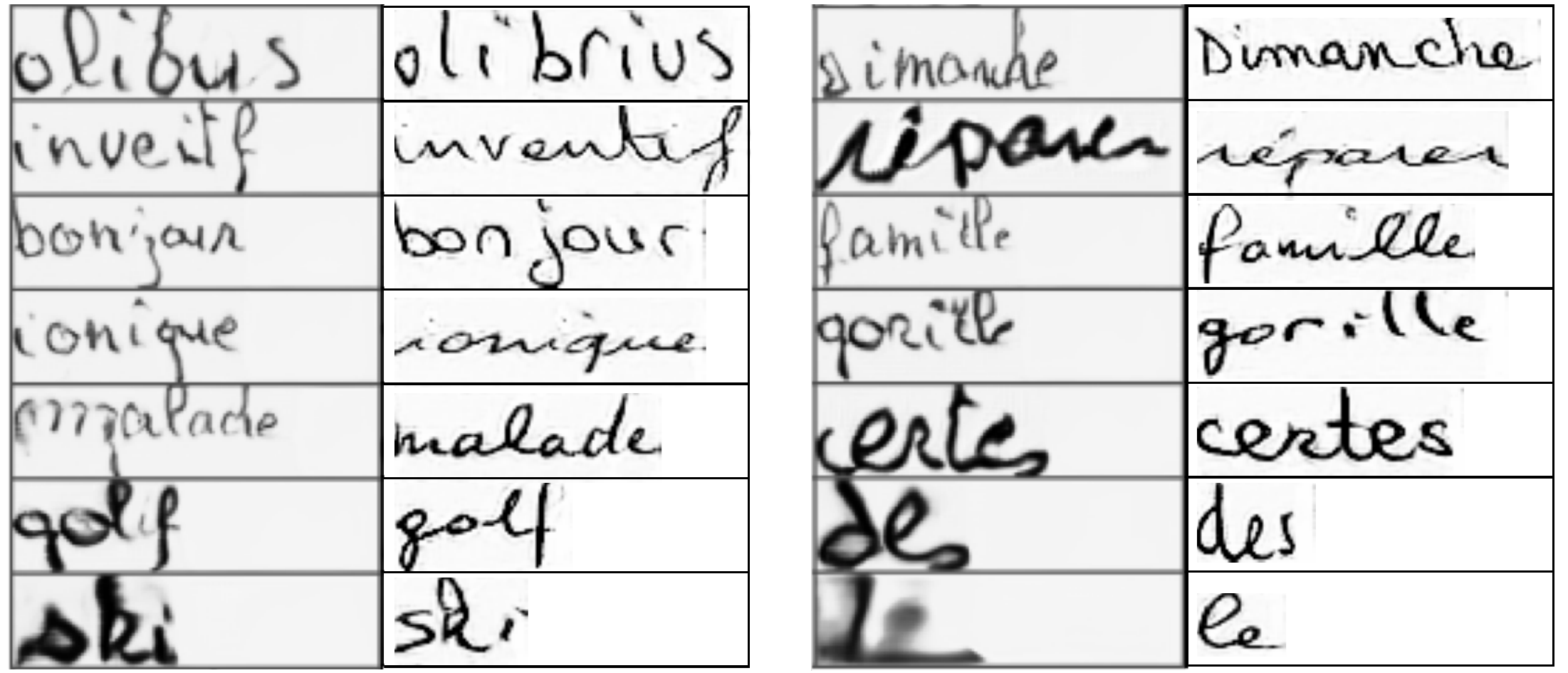}
	\caption{\label{fig:compare_rimes}
		Results of the work by Alonso et al. \cite{alonso2019adversarial} (left column) vs our results (right column) on the words: olibrius, inventif, bonjour, ionique, malade, golf, ski, Dimanche, r\'eparer, famille, gorille, certes, des, le.
	}
\end{figure}

The discriminator network $\pazocal{D}$ is inspired by BigGAN \cite{brock2018large}: 4 residual blocks followed by a linear layer with one output. To cope with varying width image generation, $\pazocal{D}$ is also fully convolutional, essentially working on horizontally overlapping image patches. The final prediction is the average of the patch predictions, which is fed into a GAN hinge-loss \cite{lim2017geometric}.

The recognition network $\pazocal{R}$ is inspired by CRNN \cite{shi2016end}. The convolutional part of the network contains six convolutional layers and five pooling layers, all with ReLU activation. Finally, a linear layer is used to output class scores for each window, which is compared to the ground truth annotation using the \emph{connectionist temporal classification} (CTC) loss
\cite{graves2006connectionist}.

Our experiments are run on a machine with one V100 GPU and 16GB of RAM.
For more details on the architecture, the reader is referred to the supplemental materials.

\subsection{Datasets and evaluation metrics} 

To evaluate our method, we use three standard benchmarks: RIMES\cite{grosicki2009icdar}, IAM \cite{marti2002iam}, and CVL~\cite{kleber2013cvl}. 
The RIMES dataset contains words from the French language, spanning about 60k images written by 1300 different authors. 
The IAM dataset contains about 100k images of words from the English language.
The dataset is divided into words written by 657 different authors. The train, test and validation set contain words written by mutually exclusive authors.
The CVL dataset consists of seven handwritten documents, out of which we use only the six that are English. 
These documents were written by about 310 participants, resulting in about 83k word crops, divided into train and test sets.

All images were resized to a fixed height of 32 pixels while maintaining the aspect ratio of the original image.
For the specific case of GAN training, and only when labels were used (supervised case), we additionally scaled the image horizontally to make each character approximately the same width as the synthetic ones, i.e. 16 pixels per character.
This was done in order to challenge the discriminator by making real samples more similar to the synthesized ones.

We evaluate our method We evaluate our method using two common gold standard metrics.
First, word error rate (WER) is the number of misread words out of the number of words in the test set. 
Second, normalized edit-distance (NED) is measured by the edit-distance between the predicted and true word normalized by the true word length.
Whenever possible, we repeat the training session five times and report the average and standard deviation thereof. 

\subsection{Comparison to Alonso el al. \cite{alonso2019adversarial}}

Since no implementation was provided, we focus on qualitative comparison to \cite{alonso2019adversarial} using images and metrics presented therein.
Figure~\ref{fig:compare_rimes} contains results shown in \cite{alonso2019adversarial} alongside results of our method on the same words.
As can be seen in the figure, our network produces images that are much clearer, especially for shorter words.
More generally, our results contain fewer artifacts, for example, the letter `m' in the fifth row, the redundant letter `i' in the sixth row and the missing `s' in the row before last.

Table~\ref{fig:compare_rimes} compares the two methods using standard metrics for GAN performance evaluation, namely \emph{Fr\'echet Inception Distance} (FID) \cite{heusel2017gans} and \emph{geometric-score} (GS) \cite{khrulkov2018geometry}.
Using a similar setting\footnote{We ran this experiment once, as opposed to \cite{alonso2019adversarial} who presented the best result over several runs} to the ones described in \cite{alonso2019adversarial}, our method shows slightly better performance on both metrics.
Note, however, that since we do not have access to the data from \cite{alonso2019adversarial}, both metrics for that method are copied from the paper, and hence cannot be used to directly compare to our results.

\subsection{Generating different styles}
\begin{figure*}[t]
	\includegraphics[width=\textwidth]{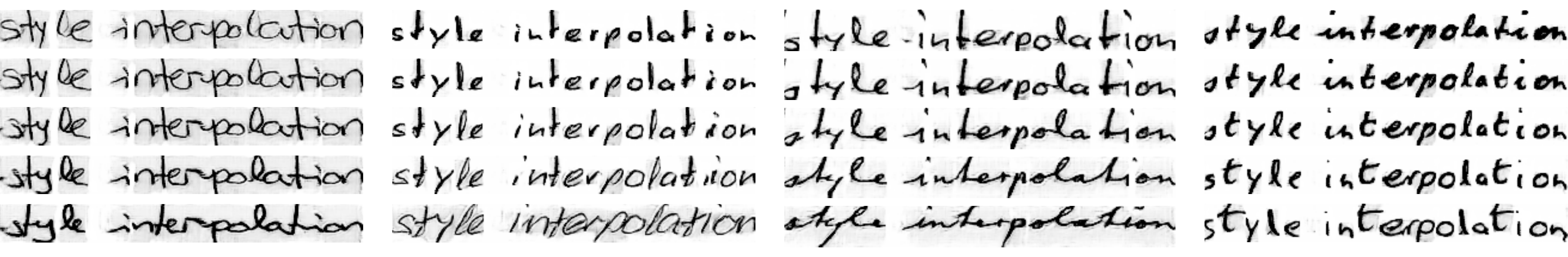}
	\caption{
		{\em Style interpolation.} 
		Each column contains an interpolation between two different styles of handwriting generated by ScrabbleGAN. 
		Note that the GAN captures the background noise typical to the IAM dataset~\cite{marti2002iam}. 
	}
\label{fig:interpolate}
\end{figure*}
We are able to generate different handwriting styles by changing the noise vector $z$ that is fed into ScrabbleGAN. Figure~\ref{fig:styles} depicts examples of selected words generated in different handwriting styles. Each row in the figure represent a different style, while each column contains a different word to synthesize. As can be seen in the figure, our network is able to generate both cursive and non-cursive text, with either a bold or thin pen stroke. 
This image provides a good example of character interaction: while all repetitions of a character start with identical filters $f_i$, each final instantiation might be different depending on the adjacent characters. 

\begin{table}[t]
\addtolength{\tabcolsep}{5pt}
\begin{tabular}{lcc}
\hline
 & FID   & GS          \\ 
\hline
\hline
Alonso el al. \cite{alonso2019adversarial}  & 23.94 & $8.58 \times 10^{-4}$ \\
ScrabbleGAN & $\pmb{23.78}$ & $\pmb{7.60 \times 10^{-4}}$ \\ 
\hline       
\end{tabular}
\caption{
	Comparison of our method to Alonso et al.\cite{alonso2019adversarial} using Fr\'echet Inception Distance and geometric-score metrics. Lower values are better.
}
\label{tab:compare}
\end{table}

Figure~\ref{fig:interpolate} shows interpolations between two different styles on the IAM dataset. In each column we chose two random noise vectors for the first and last row, and interpolated between them linearly to generate the noise vectors for the images in between.
The size of each letter, the width of the pen strokes and the connections between the letters change gradually between the two styles. 
The gray background around the letters is a property of the original IAM dataset and can be found in most of the images in the dataset. As a result, the generator also learns to generate variations of the background.

\subsection{Boosting HTR performance}

Our primary motivation to generate handwriting images is to improve the performance of an HTR framework compared to the ``vanilla'' supervised setting. 
For all experiments in this section, we use the code provided by \cite{baek2019wrong} as our HTR framework, as it contains all the improvements presented in \cite{dutta2018improving} (for which no implementation was provided), as well as some other recent advances that achieve state of the art performance on the scene text recognition problem for printed text. 
We show that training the best architecture in \cite{baek2019wrong} on the handwritten data yields performance close to state of the art on HTR, which should be challenging to improve upon. 
Specifically, our chosen HTR architecture is composed of a thin plate spline (TPS) transformation model, a ResNet backbone for extracting the visual features, a bi-directional LSTM module for sequence modeling, and an attention layer for the prediction. In all the experiments, we used the validation set to choose the best performing model, and report the performance thereof on its associated test set.

\paragraph{Train set augmentation} is arguably the most straightforward application of a generative model in this setting: by simply appending generated images to the train set, we strive to improve HTR performance in a bootstrap manner. 
Table~\ref{tab:OCR_results} shows WER and NED of the HTR network when trained on various training data agumentations on the training data, for both RIMES and IAM datasets, where each row adds versatility to the process w.r.t. its predecessor.
For each dataset, the first row shows results when using the original training data, which is the baseline for comparison. 
Next, the second row shows performance when the data is augmented with a random affine transformations.
The third row shows results using the original training data and an additional 100k synthetic handwriting image generated by ScrabbleGAN. 
The last row further fine-tunes the latter model using the original training data. 
As can be seen in the table, using the ScrabbleGAN generated samples during training leads to a significant improvement in performance compared to using only off-the-shelf affine augmentations. 
\begin{table}[h]

\addtolength{\tabcolsep}{-2pt}
\begin{tabular}{c|ccc|ll}
\hline
Set & Aug  & GAN  & Refine & WER[\%] & NED[\%]     \\ 
\hline
\hline

\multirow{4}{*}{\rotatebox{90}{RIMES}}
& $\times$		& $\times$	& - & $12.29\pm 0.15$ & $ 3.91\pm 0.08$       \\ 
& \checkmark	& $\times$	& - & $12.24\pm 0.2$ & $ 3.81\pm 0.08$       \\ 
& \checkmark	& 100k 		& $\times$ & $11.68\pm 0.29$  & $ 3.74\pm 0.10$      \\
& \checkmark	& 100k 		& \checkmark & $11.32\pm 0.31$ & $ 3.57\pm 0.13$       \\

\hline

\multirow{4}{*}{\rotatebox{90}{IAM}}
& $\times$		& $\times$	& - & $ 25.10\pm 0.49$ & $ 13.82\pm 0.35$       \\ 
& \checkmark	& $\times$	& - & $ 24.73\pm 0.53$  & $ 13.98\pm 0.93$      \\ 
& \checkmark	& 100k 		& $\times$ & $ 23.98\pm 0.4$ & $ 13.57\pm 0.24$    \\  
& \checkmark	& 100k 		& \checkmark & $ 23.61\pm 0.36$ & $ 13.42\pm 0.27$      \\ 

\hline

\end{tabular}
\caption{\emph{HTR experiments on RIMES and IAM}. 
	For each dataset we report four results with gradually increasing versatility to the dataset w.r.t. its predecessor. 
	The second column (`Aug') indicates usage of random affine augmentation in train time. 
	The third column (`GAN') indicates whether synthetic images were added to the original train set, and how many. 
	The fourth column (`Refine') indicates whether another pass of fine tuning was performed using the original data. 
	See text for more details.
}
\label{tab:OCR_results}
\end{table}

\begin{table}[t]
\addtolength{\tabcolsep}{-2pt}
\begin{tabular}{l|ll|ll}
\hline
Train data   & Style & Lex. & WER[\%] & NED[\%]     \\
\hline
\hline

IAM	(naive)		&	N/A	&	IAM	&	$39.95\pm 0.91$ & $ 19.29\pm 0.95$       \\

\hline

IAM+100K	&	CVL	&	IAM	&	$40.24\pm 0.51$ & $ 19.49\pm 0.76$       \\
IAM+100K	&	IAM	&	CVL	&	$35.98\pm 0.38$ & $ 17.27\pm 0.23$       \\
IAM+100K	&	CVL	&	CVL	&	$29.75\pm 0.67$ & $ \pmb{14.52\pm 0.51}$       \\

\hline
CVL (oracle)	&	N/A	&	CVL	&	$\pmb{22.90\pm 0.07}$ & $ 15.62\pm 0.15$       \\

\hline

\end{tabular}
\caption{
	\emph{Domain adaptation results} from the IAM dataset to the CVL dataset. 
	First row is naive approach of using a net trained on IAM. 
	Next three rows show the effect of 100k synthetic images having either CVL style, CVL lexicon or both.
	The bottom row shows the oracle performance of supervised training on the CVL train set, just for reference.
	No CVL labels were used to train HTR, except for the oracle. 
}
\label{tab:transfer_CVL}
\end{table}

\paragraph{Domain adaptation,} sometimes called transductive tra\-nsfer learning, is the process of applying a model on data from a different distribution than the one it was trained on. 
We test this task by transferring from IAM to CVL as they both use the same alphabet and are somewhat visually similar.
One naive solution for this is training a model on the IAM dataset, and testing its performance on the CVL test set. This will be our baseline for comparison.
Since ScrabbleGAN can be trained on unlabeled data, it can adapt to the style of CVL images without using the ground truth.
We synthesize data according three different flavors: using either CVL style, CVL lexicon, or both (as opposed to IAM). 
Data generated from each of these three flavors is appended to the IAM training set, as we find this helps stabilize HTR training.
Finally, we set a ``regular'' supervised training session of CVL train set, to be used as an oracle, i.e. to get a sense of how far we are from using the train labels.

Table~\ref{tab:transfer_CVL} summarizes performance over the CVL test set of all the aforementioned configurations, ranging from the naive case, through the flavors of using data from ScrabbleGAN, to the oracle.
First, we wish to emphasize the $17\%$ WER gap between the naive approach and the oracle, showing how hard it is for the selected HTR to generalize in this case. 
Second, we observe that synthesizing images with CVL style and IAM lexicon (second row) does not alter the results compared to the naive approach.
On the other hand, synthesizing images with IAM style and CVL lexicon (third row) boosts WER performance by about $5\%$.
Finally, synthesizing images with both CVL style and lexicon (fourth row) yields another $5\%$ boost in WER, with NED score that is better than the oracle.

\subsection{Gardient balancing ablation study}
Several design considerations regarding parameter selection were made during the conception of ScrabbleGAN.
We focus on two main factors: First, the effect of gradient balancing (GB) presented below, and second, the surprising effect of the architecture of the recognizer $\pazocal{R}$ which we leave to the supplementary material.

Table~\ref{tab:ablation} compares HTR results on the RIMES dataset using three different variations of gradient balancing during training:
First, we show results when no gradient balancing is used whatsoever.
Second, we apply the gradient balancing scheme suggested in \cite{alonso2019adversarial}, which is shown in Eq.~\eqref{eq:GB_alfonso}.
Finally, we show how our modified version performs for different values of the parameter $\alpha$, as described in Eq.~\eqref{eq:GB_ours}. 
For all the above options we repeat the experiment shown in the third row of Table~\ref{tab:OCR_results}, and report WER and NED scores. 
Clearly, the best results are achieved using samples synthesized from a GAN trained using our gradient balancing approach with $\alpha=1$.

\begin{table}[t]
	\addtolength{\tabcolsep}{4pt}
	\begin{tabular}{ll|ll}
		\hline
		GB Type & $\alpha $ & WER[\%] & NED[\%]     \\
		\hline
		\hline
		No GB	& -		& $12.64\pm 0.20$  & $ 4.18\pm 0.11$      \\
		\cite{alonso2019adversarial}	& 1		&  $12.83\pm 0.28$  & $ 4.21\pm 0.06$      \\
		Ours	& 0.1	& $12.28\pm 0.49$  & $ 3.95\pm 0.26$      \\
		Ours	& 1	&  $\pmb{ 11.68\pm 0.29}$  & $\pmb{ 3.74\pm 0.10}$      \\
		Ours	& 10	& $12.03\pm 0.27$  & $ 3.80\pm 0.04$      \\
		
		\hline
	\end{tabular}
	\caption{\label{tab:ablation}
		\emph{GB ablation study}, comparing HTR performance trained on different synthetic datasets. 
		Each such set was generated by a GAN with different GB scheme.
		See text for details.
		\vspace{-2ex}
	}	
\end{table}

\begin{figure}[t]
 	\begin{overpic}[width=\textwidth]{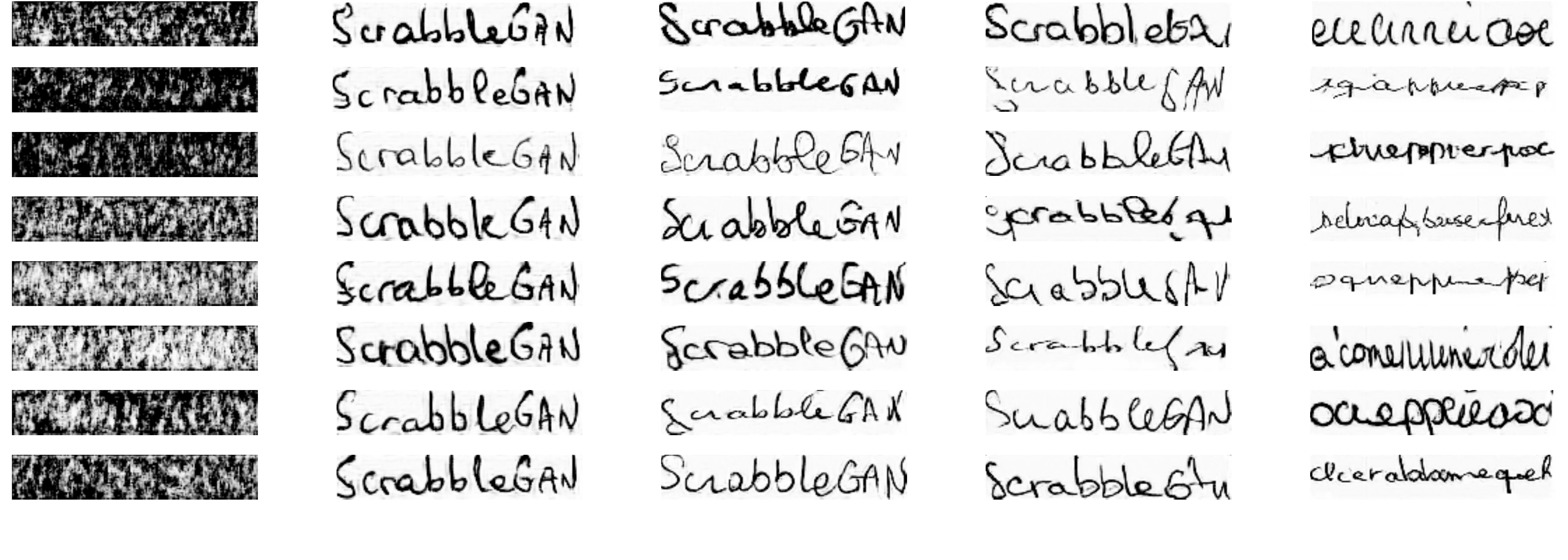}
 		\put(03,-2){$\alpha=\infty$}
 		\put(23,-2){$\alpha=10$}
 		\put(45,-2){$\alpha=1$}
 		\put(64,-2){$\alpha=0.1$}
 		\put(86,-2){$\alpha=0$}
 	\end{overpic}

    \caption{\label{fig:ablation_alpha}
		Comparison of different balancing levels between $\ell_\pazocal{D}$ and $\ell_\pazocal{R}$, the discriminator and recognizer loss terms, respectively.
		Setting $\alpha$'s value to $\infty$ or $0$ means training only with $\pazocal{R}$ or $\pazocal{D}$, respectively.
		All examples are generation of the word ``ScrabbleGAN'', where each row was generated with the same noise vector $z$.
		\vspace{-3ex}
	}
\end{figure}
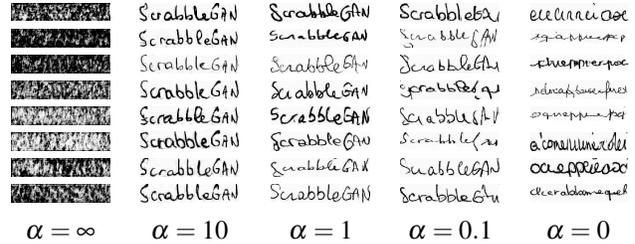

Figure~\ref{fig:ablation_alpha} further illustrates the importance of balancing between $\ell_\pazocal{R}$ and $\ell_\pazocal{D}$ and the effect of the parameter $\alpha$. 
Each column in the figure represents a different value starting from training only with $\ell_\pazocal{R}$ on the left, to training only with $\ell_\pazocal{D}$ on the right.
The same input text, ``ScrabbleGAN'', is used in all of the images and the same noise vector is used to generate each row.
As expected, using only the recognizer loss results in images which look noisy and do not contain any readable text. 
On the other hand, using only the adversarial loss results in real-looking handwriting images, but do not contain the desired text but rather gibberish.
A closer look at this column reveals that manipulating the value of $z$ changes the letter itself, rather than only the style.
From left to right, the three middle columns contain images generated by a GAN trained with $\alpha$ values of $10$, $1$, and $0.1$.
The higher the value of $\alpha$ is, the higher the weight of the $\ell_\pazocal{R}$ is.
The results using $\alpha=10$ are all readable, but contain much less variability in style.
Conversely, using $\alpha=0.1$ yields larger variability in style at the expense of the text readability, as some of the letters become unrecognizable.
The images depicted in Figure~\ref{fig:ablation_alpha} provide another explanation for the quantitative results shown in Table \ref{tab:ablation}. Training an HTR network with images generated by a GAN trained with larger $\alpha$ deteriorates the results on diverse styles, while training with images generated by a GAN trained with a smaller $\alpha$ value might lead to recognition mistakes caused by training on unclear text images.

\section{Conclusion and Future Work}
We have presented a new architecture to generate offline handwritten text images, which operates under the assumption that writing characters is a local task.
Our generator architecture draws inspiration from the game ``Scrabble''. 
Similarly to the game, each word is constructed by assembling the images generated by its characters.
The generated images are versatile in both stroke widths and general style. 
Furthermore, the overlap between the receptive fields of the different characters in the text enables the generation of cursive as well as non-cursive handwriting.
We showed that the large variability of words and styles generated, can be used to boost performance of a given HTR by enriching the training set.
Moreover, our approach allows the introduction of an unlabeled corpus, adapting to the style of the text therein.
We show that the ability to generate words from a new lexicon is beneficial when coupled with the new style.

An interesting avenue for future research is to use a generative representation learning framework such as VAE \cite{kingma2013auto} or BiGAN \cite{donahue2016adversarial,dumoulin2016adversarially}, which are more suitable for few shot learning cases like author adaptation. Additionally, disentanglement approaches may allow finer control of text style, such as cursive-ness or pen width. 

In the future, we additionally plan to address the fact that generated characters have the same receptive field width.
This is, of course, not the case for most scripts, as `i' is usually narrower than `w', for example.
One possible remedy for this is having a different width for each character filter depending on its average width in the dataset. 
Another option is to apply STN \cite{jaderberg2015spatial} as one of the layers of $\pazocal{G}$, in order to generate a similar effect.

\begin{figure}[h!]
	\includegraphics[width=\columnwidth]{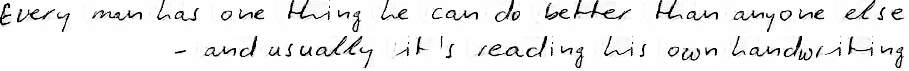}
	\flushright \tiny J. Norman Collie
\end{figure}

\clearpage
\appendix

\title{\Large{\textbf{ScrabbleGAN: Semi-Supervised Varying Length Handwritten Text Generation} \vskip5pt  Supplementary Materials}}

\author{
Sharon Fogel$^\dagger$, Hadar Averbuch-Elor$^\mathsection$, Sarel Cohen$^\dagger$, Shai Mazor$^\dagger$ and Roee Litman$^\dagger$ \\
$^\dagger$ Amazon Rekognition, Israel \qquad $^\mathsection$ Cornell Tech, Cornell University
}

\maketitle

\section{Visual Results}

\begin{figure}[h!]
	\vskip4ex

	\includegraphics[width=\columnwidth]{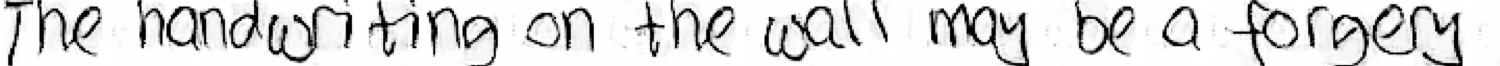}
	\flushright \scriptsize \vskip-2ex Adlai Stevenson \\

	\vskip4ex
	\includegraphics[width=\columnwidth]{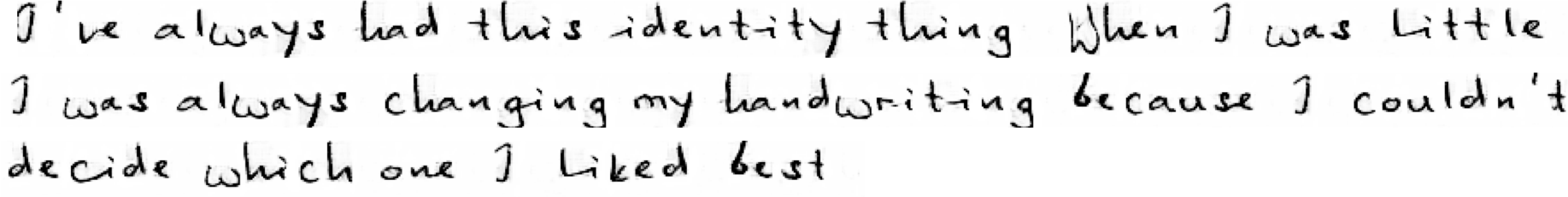}
	\flushright \scriptsize \vskip-5ex Lianne La Havas \\

	\vskip4ex	\includegraphics[width=\columnwidth]{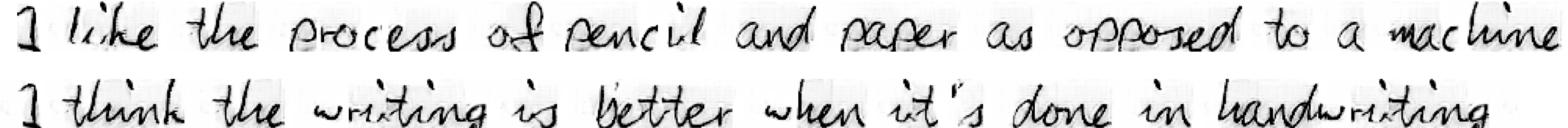}
	\flushright \scriptsize \vskip-2ex Nelson DeMille \\

	\vskip4ex	\includegraphics[width=\columnwidth]{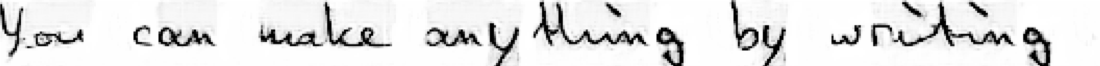}
	\flushright \scriptsize \vskip-2ex C.S Lewis \\

	\vskip4ex	\includegraphics[width=\columnwidth]{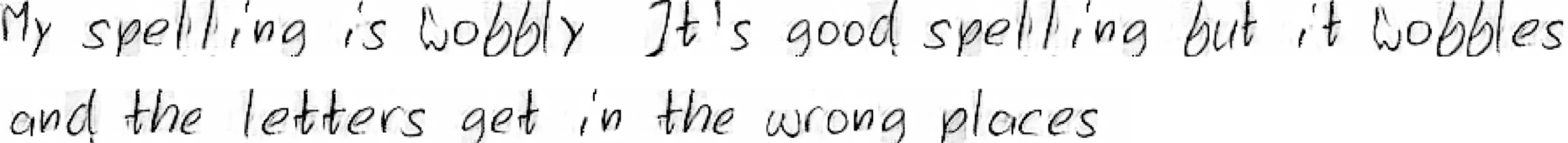}
	\flushright \scriptsize \vskip-2ex A.A. Milne, Winnie-the-Pooh \\

	\vskip4ex	\includegraphics[width=\columnwidth]{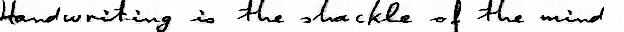}
	\flushright \scriptsize \vskip-2ex Plato \\

	\vskip4ex	\includegraphics[width=\columnwidth]{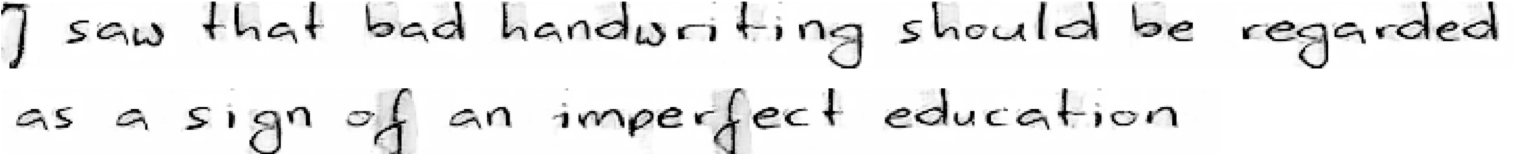}
	\flushright \scriptsize \vskip-2ex Mahatma Gandhi \\

	\vskip4ex	\includegraphics[width=\columnwidth]{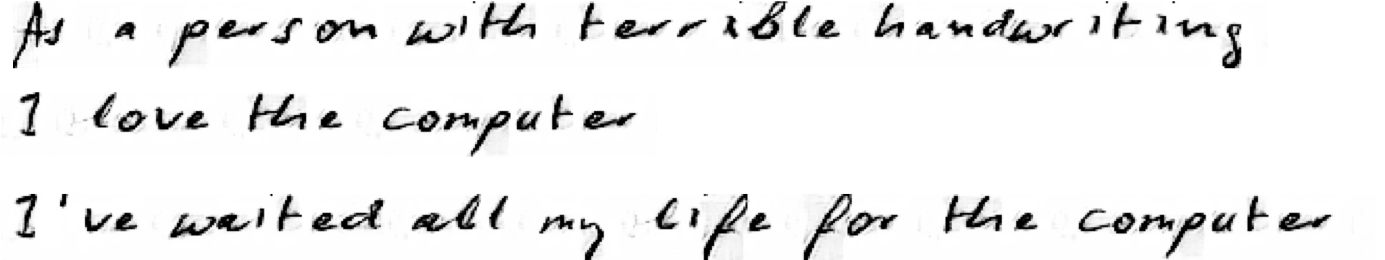}
	\flushright \scriptsize \vskip-2ex Janet Fitch \\

	\vskip4ex	\includegraphics[width=\columnwidth]{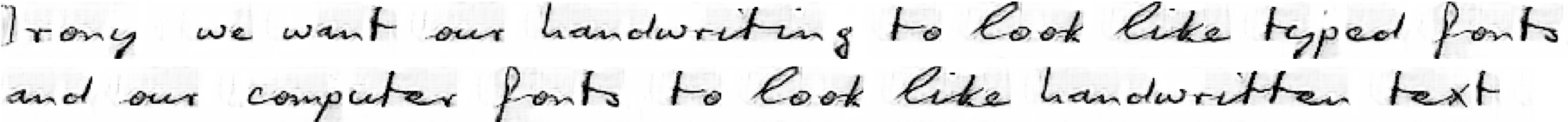}
	\flushright \scriptsize \vskip-2ex Vikrmn Corpkshetra \\
	
	\caption{
		\label{fig:supp_quotes}
		Quotes about handwriting. 
		All these examples were originally one single image, and some where split into several lines to fit one column.
	}

\end{figure}

Generating complete sentences is one application of the varying length property of ScrabbleGAN, as can be seen in the quotes about handwriting depicted in Figure~\ref{fig:supp_quotes}. 
Each quote was originally one single image, and was split into several lines to fit one column.

\section{Ablation Study}
\label{sec:ablation}

\paragraph{Ablation results.}
In Table~\ref{tab:ablation_supp} we provide results of a few more ablation experiments, justifying selection of two more components of our framework: the architecture of $\pazocal{R}$ and the way the noise vector is fed into the network. 

\begin{table}[h]
\addtolength{\tabcolsep}{0pt}
\begin{tabular}{l|ll}
\hline
Modification &  WER[\%] & NED[\%]     \\
\hline
\hline
CNN \cite{rcnn_pytorch}  	& $\pmb{ 11.68\pm 0.29}$  & $\pmb{ 3.74\pm 0.10}$      \\
\hline
CNN \cite{rcnn_pytorch} + LSTM  	& $13.80\pm 0.30$  & $ 5.30\pm 0.13$      \\
CRNN  	& 12.18$\pm 0.24$  & 3.91 $ \pm $ 0.08     \\
CRNN + LSTM  	& 12.31 $\pm $ 0.28 & 3.96$ \pm $ 0.17     \\
ResNet + LSTM + Attn  	& 12.27$\pm $ 0.34 & 3.87$ \pm $ 0.09     \\
\hline
CNN \cite{rcnn_pytorch} w/o CBN \cite{dumoulin2016learned} 	& 12.46 $\pm $ 0.30 & 4.01$ \pm $ 0.09     \\

\hline

\end{tabular}
\caption{
	\emph{Ablation results on genrator and recognizer architecture}, comparing HTR performance trained on different synthetic datasets. 
	Each such set was generated by a GAN with different generator or recognizer architecture.
	See text for details.
}
\label{tab:ablation_supp}
\end{table}

\paragraph{Recognizer architecture selection.} 
We tested several options for the recognizer network $\pazocal{R}$ to be used during GAN training.
As mentioned in Section~\ref{sec:recog} in the main paper, better HTR network will not necessarily do better for ScrabbleGAN.
Rows 3 through 5 in Table~\ref{tab:ablation_supp} present three alternatives from the code provided by \cite{baek2019wrong}.
Surprisingly, the `weakest' configuration of the three yields the best performance, despite the fact it contains no recurrent sub network.
To push this observation even further, we used a recognizer presented by \cite{rcnn_pytorch}, which contains a simple feed forward backbone of seven convolutional layers with a bidirectional LSTM on top.
We tested this architecture with- and without the LSTM module, and respectively present their performance in rows 2 and 1 of Table~\ref{tab:ablation_supp}.
Indeed, this simpler network helped the GAN generate the best images to be used for HTR training. 
Alonso el al. \cite{alonso2019adversarial} used gated CRNN as their recognizer $\pazocal{R}$, originally presented in \cite{bluche2017gated}.
Since this is very similar to the CRNN presented in \cite{baek2019wrong}, and no implementation of \cite{bluche2017gated} was provided, we chose not to include an evaluation of this specific architecture.

\paragraph{GAN noise input selection.} As we describe in Section~\ref{sec:supp_arch} below, we do not feed class data into CBN layers. This raised the option to remove these layer in favor of standard BN layers. 
As we show in the bottom row in Table~\ref{tab:ablation_supp}, doing so adds about $1\%$ to the WER score. Therefore, we opted to use CBN layers in the generator.

\section{Architecture Details}
\label{sec:supp_arch}

We now provide some more specific implementation details for the three modules that comprise ScrabbleGAN.

\paragraph{Generator and discriminator.}
\begin{table}[h]

\addtolength{\tabcolsep}{2pt}

\begin{tabular}{l|ccc}
\hline
Parameter  &  block 1 & block 2 & block 3   \\
\hline
\hline

\small{\texttt{in\_channels}}$^\dagger$		&	8	&   4   & 2     \\
\small{\texttt{out\_channels}}$^\dagger$	    &	4	&	2	& 1     \\
\small{\texttt{upsample\_width}}	    &	2	&	2   & 2     \\
\small{\texttt{upsample\_height}}	&	2	&	2   & 1     \\
\small{\texttt{resolution}}         &   8   &   16  & 16    \\
\small{\texttt{kernel1}}    	    &   3   &   3   & 3     \\
\small{\texttt{kernel2}}    	    &	3   &   3   & 1     \\

\hline

\end{tabular}
\caption{
	Generator architecture parameters used in the helper function \texttt{G\_arch} in the file \texttt{BigGAN.py}. 
	$^\dagger$ The number of input and output channels is the default parameter \texttt{ch=64} multiplied by the number of channels in the table.
}
\label{tab:G_arch}
\end{table}
\begin{table}[h]

\addtolength{\tabcolsep}{-2pt}
\begin{tabular}{l|cccc}
\hline
Parameter  &  block 1 & block 2 & block 3 & block 4  \\
\hline
\hline

\small{\texttt{in\_channels}}$^\star$		&   \small{\texttt{input\_nc}}    &	1	&   8   & 16     \\
\small{\texttt{out\_channels}}$^\dagger$	&   1    &	8	&	16	& 16     \\
\small{\texttt{downsample}}	    &   \checkmark    &	\checkmark	&	\checkmark   & $\times$     \\
\small{\texttt{resolution}}	    &   16    &	8	&	4   & 4     \\

\hline

\end{tabular}
\caption{
	Discriminator architecture parameters used in the helper function \texttt{D\_arch} in the file \texttt{BigGAN.py}. 
	$^\star$ The number of input channels in the first block is the number of channels in the image (in our case 1), and in the other blocks it is the default parameter \texttt{ch=64} multiplied by the number of channels in the table.
	$^\dagger$ The number of output channels is the default parameter \texttt{ch=64} multiplied by the number of channels in the table.
}
\label{tab:D_arch}
\end{table}

We based our implementation of $\pazocal{D}$ and $\pazocal{G}$ on the PyTorch version of BigGAN \cite{biggan_pytorch}.
The only modifications we made are in the file \texttt{BigGAN.py}. 
We changed the architecture parameter helpers \texttt{G\_arch} and \texttt{D\_arch} as described in Tables \ref{tab:G_arch} and \ref{tab:D_arch} respectively, in order to adjust the output patch to a size of $16\times 32$ pixels per character.
The code of the Generator class was changed accordingly to work with different width and height up-sampling parameters.

A few further modifications were made in the architecture of $\pazocal{G}$ to accommodate our scheme of class conditional generator. 
Unlike the original BigGAN \cite{brock2018large} where one class is used for the entire image, here different regions of the image are conditioned on different classes (characters).
Imposing this spacial condition in the first layer is easier since there is no overlap between different characters.
It is more difficult, however, to feed this information directly into the CBN layers in the following blocks, due to the receptive fields overlap. 
For this reason, we only use the noise vectors $z_2$ through $z_4$ with no class conditioning to the CBN layers.  
More details about the input to the first layer appear in the implementation details in Section~\ref{sec:implementation} in the paper.

\paragraph{Recognizer.}
For $\pazocal{R}$ we based our implementation on the RCNN implementation by \cite{rcnn_pytorch}. 
In light of the ablation presented in section \ref{sec:ablation}, we decided to remove the Bi-LSTM network.

\end{document}